\documentclass{article}

\usepackage[utf8]{inputenc} 
\usepackage[T1]{fontenc}    
\usepackage{hyperref}       
\usepackage{url}            
\usepackage{booktabs}       
\usepackage{amsfonts}       
\usepackage{nicefrac}       
\usepackage{microtype}      
\usepackage{graphicx}
\usepackage{hyphenat}
\usepackage{amssymb}
\usepackage{amsmath}
\usepackage{enumitem}
\usepackage{amsthm}
\usepackage{bm}
\usepackage[super]{nth}
\usepackage{multirow}
\usepackage{units}
\usepackage{wrapfig}
\usepackage[ruled]{algorithm2e}
\usepackage{placeins}
\usepackage{xcolor}

\newtheorem{lemma}{Lemma}

\makeatletter

\newcommand*\wthelper[2]{%
        \hbox{\dimen@\accentfontxheight#1%
                \accentfontxheight#11.3\dimen@
                $\m@th#1\widetilde{\bm{#2}}$%
                \accentfontxheight#1\dimen@
        }%
}

\newcommand*\accentfontxheight[1]{%
        \fontdimen5\ifx#1\displaystyle
                \textfont
        \else\ifx#1\textstyle
                \textfont
        \else\ifx#1\scriptstyle
                \scriptfont
        \else
                \scriptscriptfont
        \fi\fi\fi3
}
\makeatother

\newcommand{\In}{\mathbin{\in}}
\newcommand{\Eq}{\mathbin{=}}

\newcommand{\Geq}{\mathbin{\geq}}

\newcommand{\Plus}{\mathbin{+}}
\newcommand{\Minus}{\mathbin{-}}
\newcommand{\Times}{\mathbin{\times}}

\newcommand{\Bzero}{\bm{0}}

\newcommand{\Bd}{\bm{d}}
\newcommand{\Be}{\bm{e}}

\newcommand{\Bh}{\bm{h}}

\newcommand{\Bhhat}{\smash{\hat{\bm{h}}}}

\newcommand{\Bx}{\bm{x}}

\newcommand{\By}{\bm{y}}

\newcommand{\BA}{\bm{A}}

\newcommand{\BB}{\bm{B}}

\newcommand{\BBreve}{\smash{\bm{\breve{B}}}}
\newcommand{\BC}{\bm{C}}
\newcommand{\BCbar}{\overline{\bm{C}}}

\newcommand{\BD}{\bm{D}}
\newcommand{\BE}{\bm{E}}

\newcommand{\BH}{\bm{H}}
\newcommand{\BHhat}{\smash{\hat{\bm{H}}}}

\newcommand{\BHdiag}{\smash{\hat{\bm{H}}_{diag}}}
\newcommand{\BI}{\bm{I}}
\newcommand{\BP}{\bm{P}}
\newcommand{\BQ}{\bm{Q}}
\newcommand{\BR}{\bm{R}}
\newcommand{\BS}{\bm{S}}
\newcommand{\BT}{\bm{T}}
\newcommand{\BU}{\bm{U}}
\newcommand{\BV}{\bm{V}}

\newcommand{\BX}{\bm{X}}
\newcommand{\BXbar}{\overline{\bm{X}}}
\newcommand{\BY}{\bm{Y}}
\newcommand{\BYbar}{\overline{\bm{Y}}}

\newcommand{\BLambda}{\bm{\Lambda}}

\newcommand{\R}{\mathbb{R}}

\newcommand{\E}{\mathbb{E}}
\newcommand{\PSD}{\mathcal{PSD}}

\newcommand{\NN}{\mathcal{NN}}

\newcommand{\NOR}{\mathcal{NOR}}
\newcommand{\calO}{\mathcal{O}}

\newcommand{\diag}{\text{diag}}
\newcommand{\tran}{\intercal}
\makeatletter
\newcommand*{\transpose}{%
  {\mathpalette\@transpose{}}%
}
\newcommand*{\@transpose}[2]{%
  \raisebox{\depth}{$\m@th#1\intercal$}%
}
\makeatother

\usepackage[accepted]{icml2021}

\icmltitlerunning{On-the-Fly Rectification for Robust Large-Vocabulary Topic Inference}

\begin{document}

\twocolumn[
\icmltitle{On-the-Fly Rectification for Robust Large-Vocabulary Topic Inference}



\icmlsetsymbol{equal}{*}

\begin{icmlauthorlist}
\icmlauthor{Moontae Lee}{a}
\icmlauthor{Sungjun Cho}{b}
\icmlauthor{Kun Dong}{c}
\icmlauthor{David Mimno}{d}
\icmlauthor{David Bindel}{e}
\end{icmlauthorlist}

\icmlaffiliation{a}{Information and Decision Sciences, University of Illinois at Chicago, Chicago, Illinois, USA (also affiliated in Microsoft Research at Redmond, Redmond, Washington, USA)}
\icmlaffiliation{b}{Computational Science and Engineering, Georgia Tech, Atlanta, Georgia, USA}
\icmlaffiliation{c}{Applied Mathematics, Cornell University, Ithaca, New York, USA}
\icmlaffiliation{d}{Information Science, Cornell University, Ithaca, New York, USA}
\icmlaffiliation{e}{Computer Science, Cornell University, Ithaca, New York, USA}

\icmlcorrespondingauthor{Moontae Lee}{moontae@uic.edu}

\icmlkeywords{Machine Learning, ICML}

\vskip 0.3in
]



\printAffiliationsAndNotice{}

\setlength{\textfloatsep}{10pt}

\begin{abstract}

Across many data domains, co\hyp{}occurrence statistics about the joint appearance of objects are powerfully informative. By transforming unsupervised learning problems into decompositions of co\hyp{}occurrence statistics, spectral algorithms provide transparent and efficient algorithms for posterior inference such as latent topic analysis and community detection. As object vocabularies grow, however, it becomes rapidly more expensive to store and run inference algorithms on co\hyp{}occurrence statistics. Rectifying co\hyp{}occurrence, the key process to uphold model assumptions, becomes increasingly more vital in the presence of rare terms, but current techniques cannot scale to large vocabularies. We propose novel methods that simultaneously compress and rectify co\hyp{}occurrence statistics, scaling gracefully with the size of vocabulary and the dimension of latent space. We also present new algorithms learning latent variables from the compressed statistics, and verify that our methods perform comparably to previous approaches on both textual and non-textual data.
\end{abstract}

\section{Introduction}\label{sec:introduction}
\begin{figure*}[t]
 	\centering
 	\vspace{-5px}
 	\includegraphics[width=\textwidth, trim={0.2cm 5.0cm 0.2cm 2.7cm}, clip]{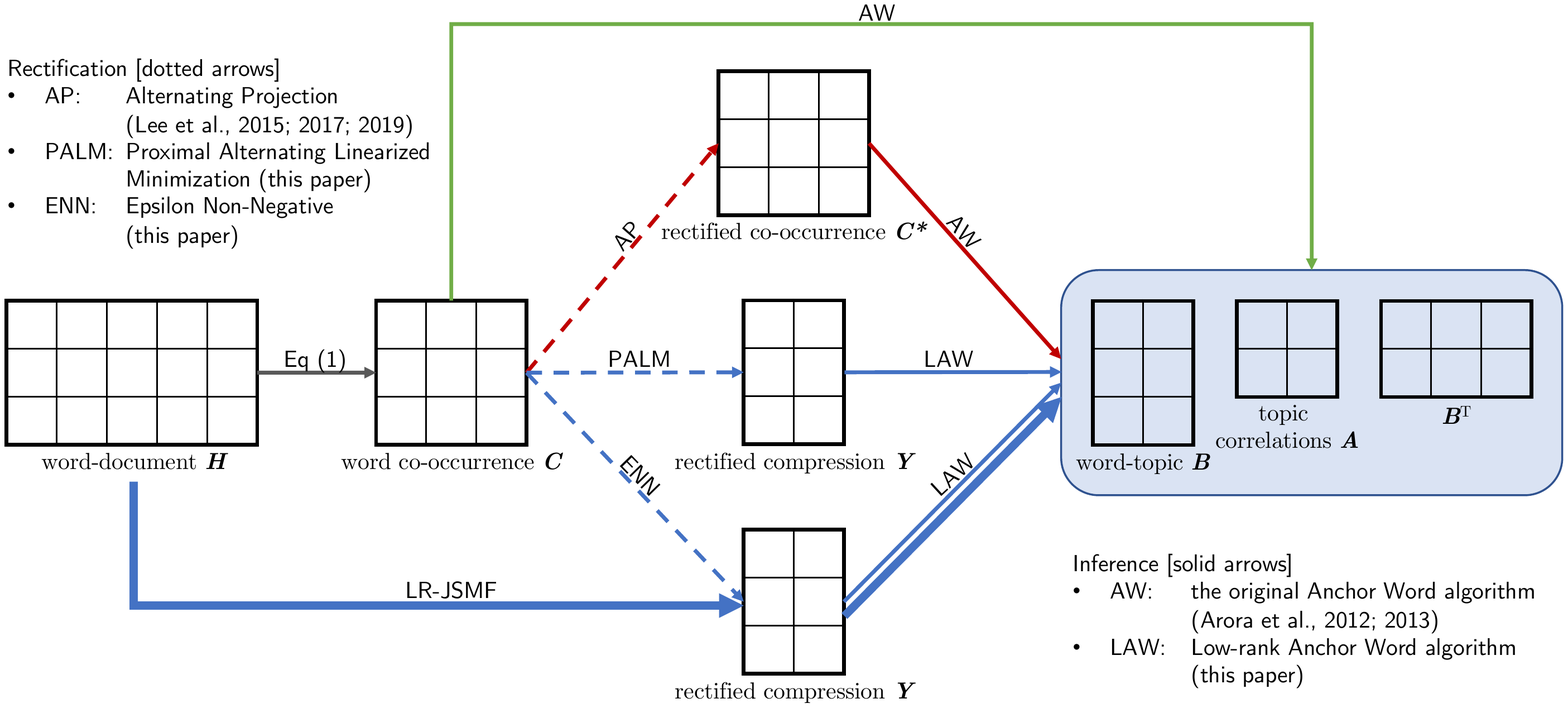}
 	\vspace{-15px}
	\caption{Overall framework. The Rectified Anchor Word algorithm ({\color{red}RAW=AP+AW}) \citep{moontae2015nips,moontae2017from,moontae2019practical} significantly improves the original Anchor Word algorithm ({\color{green}AW}) \citep{AGM,arora2013practical} by adopting AP-rectification before running the inference. Developing two novel rectification methods ({\color{blue}ENN/PALM}) and a low-rank inference algorithm ({\color{blue}LAW}), we propose a robust and scalable pipeline ({\color{blue}\textbf{LR-JSMF}}) that constructs the ENN-rectified and compressed co\hyp{}occurrence directly from a bag-of-words raw corpus possibly with large vocabulary (\nth{1} bold blue arrow), then running our LAW to efficiently learn high-quality topics and their correlations (\nth{2} bold blue arrow).}.
	\label{fig:flowchart}
\end{figure*}

Understanding the underlying geometry of noisy and complex data is a fundamental problem of unsupervised learning. Probabilistic models explain data generation processes in terms of low-dimensional latent variables. Inferring a posterior distribution for these latent variables provides us with a compact
representation for various exploratory analyses and downstream tasks \citep{bengio2013representation}. However, exact inference is often intractable due to entangled interactions between the latent variables \cite{LDA,airoldi2008mixed,erosheva2003bayesian,pritchard2000inference}. Variational inference transforms the posterior approximation into an optimization problem over simpler distributions with independent parameters \cite{JGJS,WJ,blei2017variational}, while Markov Chain Monte Carlo enables users to sample from the desired posterior distribution \cite{neal1993probabilistic,neal2011mcmc,robert2013monte}. However, these likelihood-based methods require numerous iterations without any guarantee beyond local improvement at each step \cite{kulesza2014low}.

When the data consists of collections of discrete objects, co\hyp{}occurrence statistics summarize
interactions between objects. Collaborative filtering learns low\hyp{}dimensional representations of
individual items, which are useful for recommendation systems, by explicitly decomposing the co\hyp{}occurrence of items that are jointly consumed by certain users \cite{moontae2015nips,liang2016factorization}. Word-vector models learn low\hyp{}dimensional embeddings of individual words, which encode useful linguistic biases for neural networks, by implicitly decomposing the co\hyp{}occurrence of words that appear together in contexts
\cite{pennington2014glove,levy2014neural}. If co\hyp{}occurrence provides a rich enough set of unbiased moments about an underlying generative model, spectral methods can provably learn posterior configurations from co\hyp{}occurrence information alone, without iterating through individual training examples
\cite{arora2013practical,anandkumar2012method,hsu2012spectral,AHK}.

However, two major limitations hinder users from taking advantage of spectral inference based on co\hyp{}occurrence. First, the second-order co\hyp{}occurrence matrix already grows quadratically in the number of words (e.g. objects, items, products). Pruning the vocabulary is an option, but for a retailer selling millions of long-tailed products, learning representations of only a subset of the products is inadequate. Second, inference quality is poor in real data that does not necessarily follow our generative model. Whereas
likelihood-based methods \citep{LDA,airoldi2008mixed,erosheva2003bayesian,pritchard2000inference} have an intrinsic capability to fit the data to the model despite their mismatch, sample noise can easily destroy the performance of spectral methods even if the data is synthesized from the
model \cite{kulesza2014low,moontae2015nips}.

\textbf{Rectification}, a process of projecting empirical co\hyp{}occurrence onto a manifold consistent with the posterior geometry of the model, provides a principled treatment that improves the performance of spectral inference in the face of model-data mismatch \cite{moontae2015nips}. \textit{Alternating Projection rectification (AP)} has been used to rectify the input co\hyp{}occurrence matrix to the \textit{Anchor Word algorithm (AW)}, a second-order spectral topic model \citep{moontae2015nips,moontae2017from,moontae2019practical}, but running multiple projections dominates overall inference cost even when the vocabulary is small. AP makes the co\hyp{}occurrence dense as well, exacerbating storage costs when operating on large vocabularies.

In this paper, we propose two efficient methods that simultaneously compress and rectify the co\hyp{}occurrence matrix, \textbf{Epsilon Non-Negative rectification (ENN)} and \textbf{Proximal Alternating Linearized Minimization rectification (PALM)}. We also propose the \textbf{Low-rank Anchor Word algorithm (LAW)} that learns the latent topics and their correlations only from the compressed statistics, guaranteeing the same performance as the original Anchor Word algorithm under a certain condition. Our experiments show that applying LAW after ENN learns topics of quality comparable to using AW after AP based on the full co\hyp{}occurrence. We then introduce the \textbf{Low-Rank Joint Stochastic Matrix Factorization pipeline (LR-JSMF)} that first adopts a randomized algorithm to construct a low-rank approximation of the full co\hyp{}occurrence $\BC$ directly from the raw data; then performs ENN and LAW. While PALM needs access to the full co\hyp{}occurrence, ENN can work solely with a low-rank initialization, eliminating the burden to ever construct a full co\hyp{}occurrence matrix. This new pipeline scales to large vocabularies that were previously intractable for spectral inference, and offers a 10x$\mathbf{\sim}$100x speedup over previous methods on various textual and non-textual datasets.

Note that second-order spectral topic models often rely on the \textit{separability assumption} that forces at least one anchor word for each topic. This has led to criticism in theory despite their superior performance in practice compared to probabilistic counterparts \citep{moontae2017from} and third-order tensor models \citep{moontae2019practical}. As most topic models with large vocabularies are proven separable \citep{ding2015most}, we show that our capability to process large vocabularies not only fits for modern datasets, but also alleviates the theoretical limitation. In addition, we also develop a new approach that helps better interpretation of topics by jointly reading characteristic words as well as traditional prominent words. By defining the characteristic words as the terms that are highly associated with each anchor word, we design a graph-based metric that can measure the degree of incoherence in individual topics. To the best of our knowledge, this work makes the first principled attempt to utilize anchor words for quantitative and qualitative interpretations of topics with the prominent words. Given our on-the-fly methods, users are now capable of efficiently understanding latent topics and their correlations from noisy co\hyp{}occurrence statistics within time and space complexity linear in the size of vocabulary.

\section{Foundations and Rectification}\label{sec:background}
\begin{algorithm}[t]
		\SetKwInput{KwInput}{Input}
		\SetKwInput{KwOutput}{Output}
		\DontPrintSemicolon
		\KwInput{Word co-occurrence $\BC \In \R^{N \Times N}$\\
		\hspace{29px} Number of topics $K$}
		\KwOutput{Anchor words $\BS \Eq \{s_1, ..., s_K\}$\\ 
		\hspace{35px} Latent topics $\BB \In \R^{N \Times K}$\\
		\hspace{35px} Topic correlations $\BA \In \R^{K \Times K}$}
		\Begin{
			\hspace{5px}\vspace{1px}$L_1$\hyp{}normalize the rows of $\BC$ to form $\BCbar$.\\
			\vspace{1px}\mbox{Find $\BS$ via column pivoted QR on $\BCbar^\tran$.}\;
			\mbox{Find $\BBreve$ with $\BBreve_{ki} \Eq p(\text{topic}\; k \; | \; \text{word}\; i)$ by}\\
			\mbox{\hspace{5px}solving $N$ simplex\hyp{}constrained least squares}\\
			\vspace{1px}\mbox{\hspace{5px}in parallel to minimize $\|\BCbar \Minus \BBreve^\transpose \BCbar_{\BS
			\ast}\|_F$.}\\
			\vspace{1px}Recover $\BB$ from $\BBreve$ by the Bayes' rule.\\
			Recover $\BA$ by $\BB_{\BS\ast}^{-1}\BC_{\BS\BS}\BB_{\BS\ast}^{-1}$.
		}
		\caption{Anchor Word algorithm (AW)}
	\label{alg:awa}
\end{algorithm}

Instead of using Variational inference \citep{JGJS,WJ,blei2017variational} or Markov Chain Monte Carlo  \citep{neal1993probabilistic,neal2011mcmc,robert2013monte}, our new algorithms build upon the  \textbf{Joint-Stochastic Matrix Factorization (JSMF)} \citep{moontae2015nips}. Let $\BH \In \R^{N \Times M}$ be the word\hyp{}document matrix whose $m$-th column vector $\Bh_m$ counts the occurrences of each of the $N$ words in the vocabulary in document $m$. We denote the total number of words in document $m$ by $n_m$. Given a user-specified number of topics $K$, we seek to learn a word-topic matrix $\BB \In \R^{N \Times K}$ where $\BB_{ik}$ is the conditional probability of observing word $i$ given latent topic $k$. Instead of learning $\BB$ directly from the sparse and noisy observations $\BH$, JSMF begins with constructing the joint-stochastic co\hyp{}occurrence $\BC \in \R^{N \Times N}$ as an unbiased estimator for the underlying generative topic model by
\begin{gather}
    \BC \Eq \BHhat \BHhat^\transpose \Minus \BHdiag \;\;\text{where}\;\; \Bhhat_m \Eq \frac{\Bh_m}{\sqrt{n_m(n_m \Minus 1)M}}, \nonumber\\
    \BHdiag \Eq \text{diag}\left( \sum_{m=1}^M \frac{\Bh_m}{n_m(n_m \Minus 1)M} \right).
    \label{eqn:unbiased}
\end{gather}
The Anchor Word algorithm (AW) decomposes $\BC$ into $\BB \BA \BB^\transpose$ by Algorithm \ref{alg:awa} \citep{arora2013practical,moontae2015nips}, where $\BA \In \R^{K \Times K}$ is the topic correlation matrix whose entry $\BA_{kl}$ captures the joint probability between two latent topics $k$ and $l$.\footnote{Using $\BC$ is proven to be by far more robust than using $\BH$ \citep{AGM}. Our Eq \eqref{eqn:unbiased} fixes slightly incorrect construction of $\BC$ in \citep{arora2013practical}.}
In the limit, using infinite data generated from the correct probabilistic model, $\BA$ must agree with the second-moment of the topic proportions \citep{AGM}, the Bayesian prior in the model \citep{moontae2020prior}.

As popular spectral algorithms \citep{hsu2012spectral,anandkumar2012} often fail to learn high-quality latent variables beyond synthetic data, the decomposition described above frequently fails to learn high-quality topics due to \textit{model-data mismatch} \citep{kulesza2014low}. Under the probabilistic model assumed to generate the data, $\E[\BC]$ should not only be normalized to sum to one ($\NOR$) and be entry-wise non-negative ($\NN$), but it should also be positive semi-definite with rank equal to the number of topics $K$ ($\PSD_K$) \citep{moontae2015nips}. However, the empirical $\BC$ from real data is often indefinite and full-rank due to sample noise\footnote{Rectification notably improves the quality of topics even if a finite amount of data is synthesized from the generative models.} and the unbiased construction of $\BC$ in Equation \eqref{eqn:unbiased} that penalizes all diagonal entries. The \textbf{Rectified Anchor Word algorithm (RAW)} has an additional rectification step that forces $\BC$ to enjoy the expected structures of $\E[\BC]$ before running the main factorization. The Alternating Projection rectification (AP), as given in Algorithm \ref{alg:rawa}, has been used to overcome the gap between the underlying assumptions of our models and the actual data \citep{moontae2015nips,moontae2017from,moontae2019practical,moontae2020prior}.


\begin{algorithm}[t]
		\SetKwInput{KwInput}{Input}
		\SetKwInput{KwOutput}{Output}
		\SetKwRepeat{Repeat}{repeat $with \; t=0, 1, 2, ...$}{until}
		\DontPrintSemicolon
	    \textbf{Input/Output}: Same as Algorithm \ref{alg:awa}\\
		\Begin{
			$\BC_0 \leftarrow \BC$\\
			\Repeat{converging to a certain $\BC^*$}{
			    \vspace{2px}$(\BU,\BLambda_K) \leftarrow$ Truncated-Eig($\BC_{t}, K$)\\
				\vspace{4px}$\BLambda_K^+ \leftarrow \max(\BLambda_K, 0)$\\
				\vspace{1px}$\BC^{\PSD_K} \leftarrow \BU \BLambda_K^+ \BU^\transpose$\\
				\vspace{2px}\mbox{\medmuskip=0mu $\BC^\NOR \leftarrow \BC^{\PSD_K} + \frac{1-\sum_{ij}{\BC^{\PSD_K}_{ij}}}{N^2}\Be\Be^\tran$}\\
				\vspace{3px}$\BC^\NN \leftarrow \max(\BC^\NOR, 0)$\\
				\vspace{2px}$\BC_{t+1} \leftarrow \BC^\NN$
			}
			\vspace{1px}
			\mbox{$(\BS, \BB, \BA) \leftarrow$ AW$(\BC^*, K)\;$ (Algorithm \ref{alg:awa})}
		}
		\caption{\mbox{Rectified AW algorithm (RAW)}}
	\label{alg:rawa}        
\end{algorithm}

Rectification is also important for addressing the issue of \textit{outlier bias}. Real data often exhibit rare words that are only present in a few documents. 
Co\hyp{}occurrences of
these words are inevitably sparse with large variance, but the greedy anchor selection favors choosing these outliers.
Previous work tried to bypass this problem by oversampling topics by the number of outliers under additional identifiability assumptions \citep{gillis2014fast}. This approach is not always feasible, especially for a large vocabulary 
with many rare words.
When synonyms and short documents cause undesirable sparsity to Latent Semantic Analysis \citep{landauer1998introduction}, projection onto the leading
eigenspace blurs sparse co\hyp{}occurrences. Similarly, $\PSD_K$-projection 
significantly reduces
outlier bias, and the remaining projections 
maintain the
probabilistic structures of $\BC$, which then allow users to recover $\BB$ and $\BA$ in Algorithm \ref{alg:awa}.

\begin{algorithm}[t]
	\SetKwInput{KwInput}{Input}
	\SetKwInput{KwOutput}{Output}
	\SetKwRepeat{Repeat}{repeat $with \; t=0, 1, 2, ...$}{until}
	\DontPrintSemicolon
	\KwInput{Word co-occurrence $\BC \In \R^{N \Times N}$\\
	\hspace{29px} Number of topics $K$}
	\KwOutput{Rectified compression $\BY \In \R^{N \Times K}$}
	\Begin{
		\hspace{5px}\vspace{2px}$\BE \leftarrow \bm{0}\in\R^{N\times N}\;\;$ (sparse format)\\
		$\BC^{op}: \Bx \rightarrow \BC \Bx\;\;$ (Implicit operator)\\
		\Repeat{$\BE$ converges}{\vspace{2px}
			\hspace{5px}\vspace{2px}$(\BU,\BLambda_K) \leftarrow$ Truncated-Eig($\BC^{op}, K$)\\
			\vspace{2px}$\BLambda_K^+ \leftarrow \max(\BLambda_K, 0)$\;
			\vspace{2px}$\BY \leftarrow \BU(\BLambda_K^+)^{1/2}$\\
			\vspace{2px}$\BE_{ij} \leftarrow \max(-\BY_{i\ast}\BY_{j\ast}^\transpose, 0)$\\
			\vspace{2px}$r \leftarrow (1 - \|\BY^\transpose\Be\|_2^2-\sum_{ij}\BE_{ij})/N^2$\\
			\mbox{$\BC^{op}:\Bx \rightarrow \BY(\BY^\transpose\Bx) + \BE\Bx + r(\Be^\tran\Bx)\Be$\;}
			}
		}
		\caption{ENN-rectification (ENN)}
	\label{alg:enn}        
\end{algorithm}

Handling a \textit{large vocabulary} is another major challenge for spectral methods. Even if we limit our focus only to second-order models, the space complexity of RAW is already $\calO(N^2)$.
We are unable to exploit the high sparsity of $\BC$ as a single iteration of AP makes $\BC$ significantly denser. The three projections in AP and the rest of AW in Algorithm \ref{alg:awa} have time
complexities of $\calO(N^2 K)$, $\calO(N^2)$, $\calO(N^2)$ and $\calO(N^2 K)$, respectively.  
On the other hand, the \textit{separability assumption} is crucial for second-order models. While a line of research has tried to relax this assumption \citep{bansal2014,huang2016}, it is formally shown that most topic models are indeed separable if their vocabulary sizes are sufficiently larger than the number of topics \citep{ding2015most}, again emphasizing the urgency of an approach with better time and space scaling in the vocabulary size.

\section{Simultaneous Rectification \& Compression}\label{sec:lrrc}
The rectified co-occurrence $\BC^*$ in Section \ref{sec:background} must be of rank $K$ and positive semidefinite, hinting at an opportunity to represent it as $\BY \BY^\transpose$ for some $\BY \In \BR^{N \Times K}$. One idea for achieving this structure is to use a low-rank representation $\BC_t \Eq \BY_t \BY_t^\transpose$ throughout the rectification in Algorithm \ref{alg:rawa}. Another way to obtain this structure is to directly minimize $\|\BC \Minus \BY \BY^\transpose\|_F$ with the necessary constraints. Note that random projections cannot preserve algebraic properties of co\hyp{}occurrence other than L2-norms. Using such projections does not have any rectification effect, decreasing the performance as reported in \citep{moontae2014emnlp}. In this section, therefore, we propose two novel compression-plus-rectification algorithms based on these two ideas.

\subsection{ENN: Epsilon Non-Negative Rectification}

The Alternating Projection rectification (AP) in Algorithm~\ref{alg:rawa} produces low-rank intermediate matrices from the positive semi-definite projection ($\PSD_K$) and the normalization projection ($\NOR$), but the final projection to enforce elementwise non-negativity ($\NN$) destroys this low-rank structure. However, the $\NN$ projection significantly changes only a few elements; that is, the output of the $\NN$ projection at step $t$ is nearly rank $K \Plus 1$ plus a sparse correction $\BE_t$. The Epsilon Non-Negative rectification (ENN) in Algorithm~\ref{alg:enn} has the same structure as Algorithm~\ref{alg:rawa}, but with a key difference that it returns a sparse-plus-low-rank representation of the $\NN$ projection rather than a dense representation. Matrix-vector products with this sparse-plus-low-rank representation require $\calO(NK + \operatorname{nnz}(\BE_t))$ time, and $\calO(K)$ such matrix-vector products can be used in a Lanczos eigensolver to compute the truncated eigendecomposition at the start of the next iteration.

\begin{algorithm}[t]
		\SetKwInput{KwInput}{Input}
		\SetKwInput{KwOutput}{Output}
		\SetKwRepeat{Repeat}{repeat $with \; t=0, 1, 2, ...$}{until}
		\DontPrintSemicolon
		\KwInput{Word co-occurrence $\BC \In \R^{N \Times N}$\\
		\hspace{29px} Number of topics $K$}
		\KwOutput{Rectified compression $\BY \In \R^{N \Times K}$}
		\Begin{
			\hspace{5px}$(\BU, \BLambda_K) \leftarrow$ Truncated-Eig$(\BC, K)$\\
			$(\BX_0, \BY_0) \leftarrow (\BU \sqrt{\BLambda_K}, \BU \sqrt{\BLambda_K})$\\
            \Repeat{$\BY$ converges}{
                \hspace{6px}$c_t \leftarrow \gamma L_1(\BY_{t})$\\
                \vspace{1px}$\BX_{t+1}' \leftarrow \BX_{t} - (1/c_t)\nabla_{\BX} J(\BX_{t}, \BY_{t})$\\
                \vspace{1px}$\BX_{t+1} \leftarrow \max(\BX_{t+1}', 0)$\\
                \vspace{1px}$d_t \leftarrow \gamma L_2(\BX_{t+1})$\\
                \vspace{1px}$\BY_{t+1}' \leftarrow \BY_{t} - (1/d_t)\nabla_{\BY} J(\BX_{t+1}, \BY_{t})$\\
                $\BY_{t+1} \leftarrow \max(\BY_{t+1}', 0)$
            }
		}
		\caption{PALM-rectification (PALM)}
	    \label{alg:palm}      
\end{algorithm}

Maintaining a sparse correction matrix $\BE_t$ at each step lets the ENN approach avoid storage overheads of the original AP. To overcome the quadratic time cost at each iteration, though, we need to avoid explicitly computing every element of the intermediate $\BY \BY^\transpose$ in the course of the $\NN$ projection. However, we can bound the magnitude of elements of $\BY\BY^\transpose$ by the Cauchy-Schwartz inequality: $|\BC_{ij}| \leq \|\By_i\|_2 \|\By_j\|_2$ where $\By_i$ and $\By_j$ denote columns of $\BY^\transpose$. Let $I$ denote the index set \smash{$\{i :\|\By_i\|^2_2 > \epsilon\} \subseteq [N]$} for given $\epsilon$; then every large entry of $\BC$ belongs to either $\BY_{I \ast}\BY^\transpose$ or $\BY(\BY^\transpose)_{\ast I}$. As $\BC$ is symmetric, checking the negative entries in $\BY_{I *}\BY^\transpose$ is sufficient to find a symmetric correction $\BE$ that guarantees $\BY\BY^\transpose \Plus \BE \Geq -\epsilon$. We refer to this property as \textit{Epsilon Non-Negativity}: $\epsilon$ balances the trade-off between the effect of leaving small negative entries versus increasing the size of $I$ to look up. Instead of fixing $\epsilon$ and finding $I = \{i: \|\bm{y}_i\|_2^2 > \epsilon\}$, we set $\bm{Y}_{I*}$ to be $\mathcal{O}(K)$ rows of $\bm{Y}$ with the largest 2-norms based on the common sampling complexity of a suitable set of rows for a near\hyp{}optimal rank-$K$ approximation in subset selection.

\begin{algorithm}[t]
		\SetKwInput{KwInput}{Input}
		\SetKwInput{KwOutput}{Output}
		\DontPrintSemicolon
        \KwInput{Rectified compression $\BY \In \R^{N \Times K}$}
		\KwOutput{Anchor words $\BS \Eq \{s_1, ..., s_K\}$\\ 
		\hspace{35px} Latent topics $\BB \In \R^{N \Times K}$\\
		\hspace{35px} Topic correlations $\BA \In \R^{K \Times K}$}
		\Begin{
			\hspace{5px}Calculate row sums $\Bd = \BY(\BY^\transpose\Be)$.\\
			\mbox{Compute QR decomposition of $\BY = \BQ\BR$.}\\
			\mbox{Form $\BYbar = \diag(\Bd)^{-1}\BY$ and $\BX = \BYbar\BR^\transpose$.}\\
			\mbox{Select $\BS$ using column pivoted QR on $\BX^\transpose$.}\\
			Solve $n$ simplex-constrained least square problems\\
			\mbox{\hspace{5px} to minimize $\|\BX-\BBreve\BX_{\BS\ast}\|_F$.}\\
			Recover $\BB$ from $\BBreve$ using Bayes' rule.\\
			Recover $\BA = \BB_{\BS \ast}^{-1}\BY^{}_{\BS\ast}\BY_{\BS\ast}^\transpose\BB_{\BS \ast}^{-1}$.
		}
		\caption{Low-rank AW (LAW)}
	    \label{alg:law}      
\end{algorithm}

\subsection{PALM: Proximal Alternating Linearized Minimization Rectification}

To avoid small negative entries, we investigate another rectified compression algorithm that directly minimizes $\|\BC \Minus \BY \BY^\transpose\|_F$ subject to the stronger $\NN$-constraint $\BY \Geq 0$ and the usual $\NOR$-constraint $\|\BY^\transpose \Be\|_2 \Eq 1$. Concretely, 
\vspace*{-7px}
\begin{gather}
    \text{minimize} \;\; J(\BX, \BY) := \frac{1}{2} \| \BC - \BX \BY^\transpose \|_F^2 + \frac{s}{2} \| \BX - \BY \|_F^2 \nonumber\\ 
    \text{subject to} \;\; \BX \geq 0, \BY \geq 0. \label{eqn:palm}
\end{gather}
$\PSD_K$- and $\NOR$-constraints are implicitly satisfied by jointly minimizing the two terms in the objective function $J$, whereas $\NN$-constraint is explicit in the formulation. Thus we can apply the Proximal Alternating Linearized Minimization \citep{bolte2014proximal} for learning $\BY$ given $\BC$;
the relevant proximal operator is $\NN$ projection of $\BY$, which takes $\calO(NK)$ time at most. Note that $J$ is semi-algebraic (as it is a real polynomial) with two partial derivatives: $\nabla_{\BX}J \Eq (\BX \BY^\transpose \Minus \BC)\BY \Plus s(\BX \Minus \BY)$ and $\nabla_{\BY}J \Eq (\BY \BX^\transpose \Minus \BC)\BX \Plus s(\BY \Minus \BX)$. Thus the following lemma guarantees global convergence.\footnote{One could further improve the performance of PALM-rectification by using an inertial version like iPALM in \citep{pock2016inertial}. As PALM is not our main rectification method for the scalable pipeline, we leave this as a space for future work.}
\begin{lemma}
    For any fixed $\BY$, $\nabla_{\BX}J(\BX, \BY)$ is globally Lipschitz continuous with the moduli $L_1(\BY) \Eq \| \BY^\transpose \BY \Plus s\BI_K \|_2$. So is $\nabla_{\BY}J(\BX, \BY)$ given any fixed $\BX$ with $L_2(\BX) \Eq \| \BX^\transpose \BX \Plus s\BI_K \|_2$.
\end{lemma}
\vspace{-10px}
\begin{proof}
$\| \nabla_{\BX}J(\BX, \BY) \Minus \nabla_{\BX}J(\BX', \BY) \|_F \Eq \| (\BY^\transpose \BY \Plus s\BI_K)(\BX \Minus \BX')\|_F \leq \|\BY^\transpose \BY \Plus s\BI_k\|_2 \cdot \|\BX \Minus \BX'\|_F$. The proof is symmetric for the other case with $L_2(\BX) \Eq \| \BX^\transpose \BX \Plus s\BI_K \|_2$.
\end{proof}
\vspace{-5px}
Algorithm \ref{alg:palm} shows PALM with adaptive learning rate control based on our 2-norm Lipschitz modulus.


\section{Low-rank Anchor Word Algorithm and Scalable Pipeline}\label{sec:law}

ENN and PALM both output a compressed co\hyp{}occurrence matrix $\BY$ with $\BC \approx \BY \BY^\transpose$. In this section, we present the \textbf{Low-rank Anchor Word algorithm (LAW)} to find anchor words in $\calO(NK^2)$ (rather than $\calO(N^2K)$) directly from $\BY$. Note that LAW applies whenever $\BC$ is in a low-rank representation, which does not have to be derived from our methods. In addition, LAW performs an exact inference if $\BY \BY^\transpose \geq \Bzero$ but robust in practice when it has small negative entries as in the case with ENN.

The first step is to $L_1$\hyp{}normalize the rows of $\BC$. Given $\BC \Geq 0$, the $L_1$\hyp{}norm of each row is simply the sum of all its entries, so we can calculate the row norms by $\Bd \Eq \BY(\BY^\transpose\Be)$. To obtain the normalized $\BC$, we simply scale the rows of $\BY$, and \smash{$\BCbar \Eq (\diag(\Bd)^{-1}\BY)\BY^\transpose \Eq \BYbar\BY^\transpose$}. These steps cost $\calO(NK)$. Next, we need to apply column pivoted QR to \smash{$\BCbar^\tran$} in order to identify the pivots as our anchor words $\BS$. By taking the QR decomposition $\BY \Eq \BQ\BR$, \smash{$\BCbar^\tran$} can be further transformed into \smash{$\BQ\BR\BYbar^\tran$}. Notice that \smash{$\BCbar^\tran$} is an orthogonal embedding of \smash{$\BX^\transpose = \BR\BYbar^\tran$} onto a higher-dimensional space, which preserves the column $L_2$-norms. Lemma \ref{lem:qrequiv} shows that column pivoted QR on \smash{$\BCbar^\tran$} and on \smash{$\BR\BYbar^\tran$} are equivalent, which allows us to lower the computation cost from $\calO(N^2K)$ to $\calO(NK^2)$. 

\begin{algorithm}[t]
		\SetKwInput{KwInput}{Input}
		\SetKwInput{KwOutput}{Output}
		\DontPrintSemicolon
        \KwInput{Raw word-document $\BH \In \R^{N \Times M}$}
		\KwOutput{Anchor words $\BS \Eq \{s_1, ..., s_K\}$\\ 
		\hspace{35px} Latent topics $\BB \In \R^{N \Times K}$\\
		\hspace{35px} Topic correlations $\BA \In \R^{K \Times K}$}
		\Begin{\vspace{3px}
		    \hspace{5px}\vspace{4px}Get $\BHhat, \BHhat_{diag}$ from $\BH$ by \eqref{eqn:unbiased}.\\
			\vspace{4px}$\BC_{op}: \Bx\rightarrow\BHhat (\BHhat^\transpose\Bx) - \BHdiag\Bx$\\
			\vspace{4px}$(\BV,\BD) \leftarrow$ Randomized-Eig($\BC_{op}, K$)\\
			\vspace{4px}Initialize ENN with $\BV, \BD$.\\
			\vspace{3px}$\BY\leftarrow$ ENN-rectification\\
			\vspace{3px}$(\BS, \BB, \BA) \gets\; $LAW$(\BY)\;$ (Algorithm \ref{alg:law})
		}
		\caption{Low-rank JSMF (LR-JSMF)}
	\label{alg:lr-jsmf}        
\end{algorithm}

\begin{lemma}\label{lem:qrequiv}
  Let $\BS$ be the set of pivots that have been selected by column pivoted QR on \smash{$\BCbar^\tran \Eq \BQ\BX^\transpose$}. Given the QR decomposition, \smash{$\BXbar_{\BS\ast}^\tran \Eq \BP\BT$}, then \smash{$\BCbar_{\BS\ast}^\tran \Eq (\BQ\BP)\BT$} is the corresponding QR decomposition for the columns of $\BCbar$. For any remaining row $i \In [N]\setminus \BS \; \text{where} \; [N] \Eq \{1, 2, ..., N\}$, \begin{equation}
      \|(\BI-\BP\BP^\transpose)\BXbar_{i\ast}^\tran\|_2 = \|(\BI-(\BQ\BP)(\BQ\BP)^\transpose)\BCbar_{i\ast}^\tran\|_2
  \end{equation}
  Therefore, the next column pivot is identical for \smash{$\BCbar^\tran$} and \smash{$\BXbar^\tran$}. By induction, column pivoted QR on \smash{$\BCbar^\tran$} and \smash{$\BXbar^\tran$} return the same pivots.
\end{lemma}

\begin{figure*}[t]
 	\centering
 	\vspace{-5px}
 	\includegraphics[width=\textwidth, trim={16.0cm 4.5cm 11.0cm 4.5cm}, clip]{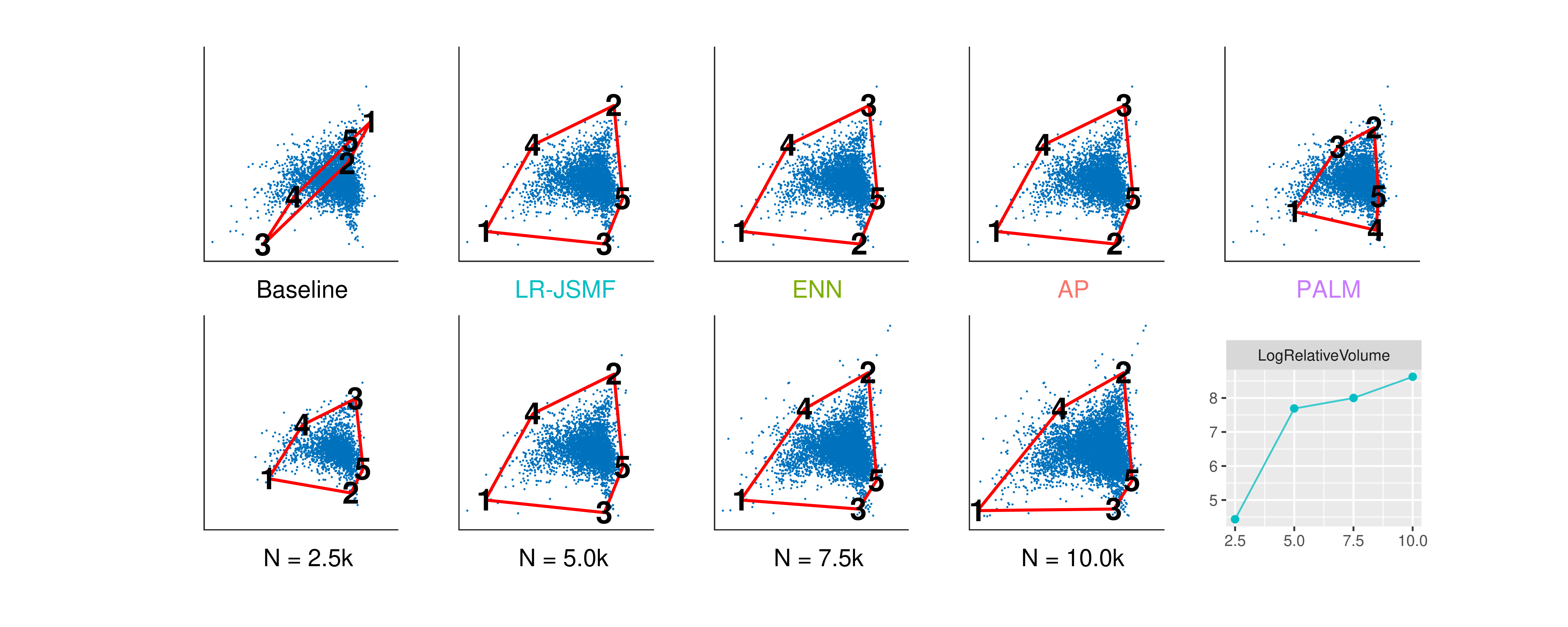}
	\vspace{-15px}
	\caption{2D visualizations of co-occurrence spaces in NeurIPS. Blue dots are words. Each vertex on the convex hulls corresponds to the $k$-th anchor word. First row shows that ENN and LR-JSMF find same anchors as AP \citep{moontae2015nips}. PALM is off but outperforms Baseline, which is AW without any rectification \citep{arora2013practical}. Second row shows growing anchor convex hull volumes relative to random convex hulls when expanding vocabularies.}
	\label{fig:cspace}
\end{figure*}

\begin{proof}
  Because both $\BQ$ and $\BP$ have orthonormal columns,
  \begin{equation*}
    (\BQ\BP)^\transpose(\BQ\BP) = \BP^\transpose(\BQ^\transpose\BQ)\BP = \BP^\transpose\BP = \BI
  \end{equation*}
  Thus, $\BQ\BP$ and $\BT$ form the QR decomposition of $\BCbar_{\BS\ast}^\tran$. The residual of a remaining column $i\in[N]\setminus \BS$ is $(\BI-(\BQ\BP)(\BQ\BP)^\transpose)\BCbar_{i\ast}^\tran$ and $(\BI-\BP\BP^\transpose)\BXbar_{i\ast}^\tran$ for $\BCbar^\tran$ and $\BXbar^\tran$, respectively. Simplify the former gives us
  \begin{align*}
    &(\BI-(\BQ\BP)(\BQ\BP)^\transpose)\BCbar_{i\ast}^\tran\\
    =& (\BI-(\BQ\BP)(\BQ\BP)^\transpose)\BQ\BXbar_{i\ast}^\tran \\
    =& \BQ\BXbar_{i\ast}^\tran - \BQ\BP\BP^\transpose\BQ^\transpose\BQ\BXbar_{i\ast}^\tran\\
    =& \BQ(\BI-\BP\BP^\transpose)\BXbar_{i\ast}^\tran
  \end{align*}
  Finally,
  \begin{align}
    \nonumber &\|(\BI-(\BQ\BP)(\BQ\BP)^\transpose)\BCbar_{i\ast}^\tran\|_2^2 \\ 
    \nonumber =& \BXbar_{i\ast}(\BI-\BP\BP^\transpose)\BQ^\transpose\BQ(\BI-\BP\BP^\transpose)\BXbar_{i\ast}^\tran\\ 
    =& \|(\BI-\BP\BP^\transpose)\BXbar_{i\ast}^\tran\|_2^2 \label{eqn:normpres}
  \end{align}
  Because the next pivot is selected as the column whose residual has the largest $L_2$\hyp{}norm, Eq. \ref{eqn:normpres} indicates that the same pivot will be selected for $\BCbar^\tran$ and $\BXbar^\tran$. Inductively, the anchors $\BS$ recovered by column pivoted QR on those matrices are equivalent.
\end{proof}
\vspace{-5px}

Following the recovery of $\BS$, AW solves $N$ independent simplex-constrained least square problems \smash{$\|\BCbar_{i\ast} - \BBreve_{i\ast}^\transpose\BCbar_{\BS\ast}\|_2$}. Again we can leverage the $L_2$-norm preserving property, 
\begin{align}\label{eqn:lrls}
  \|\BCbar_{i\ast} - \BBreve_{i\ast}^\transpose\BCbar_{\BS\ast}\|_2 &= \|\BX_{i\ast}\BQ^\transpose - \BBreve_{i\ast}^\transpose\BX_{\BS\ast}\BQ^\transpose\|_2 \nonumber \\
  &= \|\BX_{i\ast} - \BBreve_{i\ast}^\transpose\BX_{\BS\ast}\|_2
\end{align}
and reduce the dimension of the least\hyp{}square problems from $N$ to $K$, thereby reducing the complexity from $\calO(N^2K)$ to $\calO(NK^2)$. The remaining part of the algorithm follows exactly as AW.

\begin{figure*}[t]
 	\centering
	\includegraphics[width=\textwidth, trim={0.0cm 0.0cm 0.1cm 0.0cm}, clip]{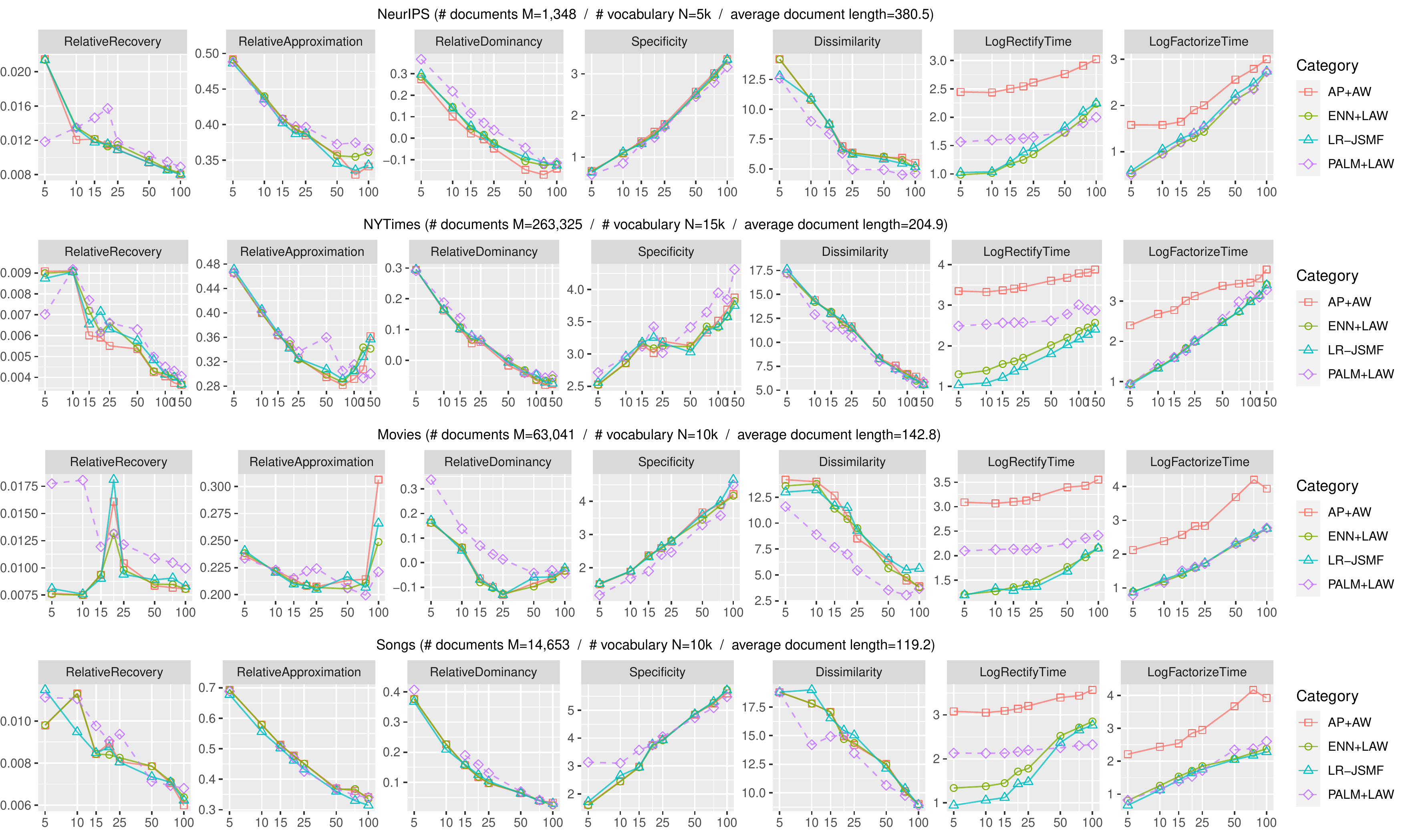}
	\vspace{-17px}
	\caption{Experiment on four datasets. ENN and LR-JSMF agree with AP, while PALM has slight inconsistency. The basic dataset statistics is above each row. Runtimes are in $\log_{10}$ seconds. Note that ENN and LR-JSMF are almost two orders of magnitude faster than AP. The $x$-axis indicates the number of topics $K$. In $y$-axes, lower is better except for Specificity and Dissimilarity.} 
	\label{fig:results-topics}
\end{figure*}

\vspace{-10px}
\paragraph{Low\hyp{}rank Joint Stochastic Matrix Factorization (LR-JSMF)} 
We complete our scalable framework of processing co\hyp{}occurrence statistics by introducing a direct initialization method for ENN from the raw word\hyp{}document data. This allows us to avoid creating and storing $\BC$, which is a burden of memory when $N$ becomes large. In Algorithm \ref{alg:enn}, $\BC$ only appears in the initial truncated eigendecomposition, after which we maintain the compressed operator $\BC_{op}$ independent of it. On the other hand, we just need the matrix\hyp{}vector multiplication by $\BC$ for iterative methods in initialization. Using the generative formula in Equation (\ref{eqn:unbiased}), we are able to implicitly apply $\BC$ to vectors as an outer\hyp{}product plus diagonal operator in terms of $\BH$, at $\calO(NMK)$ computation cost.

Note that even when $M > N$, using $\BH$ is still more efficient than using $\BC$ due to higher sparsity of $\BH$.
To further reduce the number of times the operator is applied, we adopt the one-pass randomized eigendecomposition by Halko et al. \citep{halko2011finding}. 
This technique enables initialization with a single pass over the dataset, without concurrently storing the entire $\BH$ in memory. 
A limitation is when the number of topics is large and the gap between the $K$\hyp{}th eigenvalue and the ones below is small, we will have to incorporate a few power iterations for refinement, as suggested by the original paper. This will result in a multi\hyp{}pass method, but still far more efficient on large vocabularies and parallelization-friendly.

\section{Experimental Results}

A good factorization should be accurate, meaningful, and fast. The first row of Figure \ref{fig:cspace} illustrates that using ENN with LAW or LR-JSMF pipeline directly with the raw data correctly recover the anchor convex hulls. The second row shows that the volumes of the anchor convex hulls relative to the volumes of the random convex hulls grow over increasing vocabularies.\footnote{We build a set of random convex hulls by uniformly sampling each row of $\BCbar$ from the corresponding simplex, then finding the anchors by the same column-pivoted QR on $\BCbar^\transpose$.} In the next two series of experiments, we demonstrate that our simultaneous rectification and compression maintains model quality while running in a fraction of the space and time needed for the original JSMF framework. The code is publicly available.\footnote{\url{https://github.com/moontae/JSMF}}

For the first series of experiments, we measure the accuracy of each rectification component as well as the entire pipeline of LR-JSMF. For thorough comparisons, we construct the full co-occurrence $\BC$ from each of our datasets $\BH$ by \eqref{eqn:unbiased}, and we produce the rectified $\BC_{AP}$ by running AP on $\BC$. Next we compress $\BC$ into $\BY_{ENN}$ and $\BY_{PALM}$ by running ENN (with $|I|\Eq 10K \Plus 1000$) and PALM (with $s\Eq1e^{-4}$) until convergence. To test the complete low-rank pipeline, we also construct $(\BV, \BD)$ from the raw data $\BH$ by the randomized eigendecomposition in Algorithm \ref{alg:lr-jsmf}, learning the rectified and compressed statistics \smash{$\BY_{LR-JSMF}$} by running ENN initialized with \smash{$\BV \sqrt{\BD}$}. Then we run AW on $\BC_{AP}$ and LAW on each of $\BY_{ENN}$, $\BY_{PALM}$, and $\BY_{LR-JSMF}$.

The goal of rectification is to ensure robust spectral inference to data that does not follow our modeling assumptions, so we evaluate on real data: two standard textual datasets from the UCI Machine Learning repository (NeurIPS papers and New York Times articles) as well as two non-textual datasets (Movies from Movielens 10M star-ratings and Songs from Yes.com complete playlists) previously used to show the performance of JSMF with AP in \citep{moontae2015nips}. We apply identical vocabulary curation with \citep{moontae2015nips} for fair comparisons. Data statistics are given in each figure.


\begin{figure*}[t]
 	\centering
 	\includegraphics[width=\textwidth, trim={0.0cm 0.0cm 0.1cm 0.0cm}, clip]{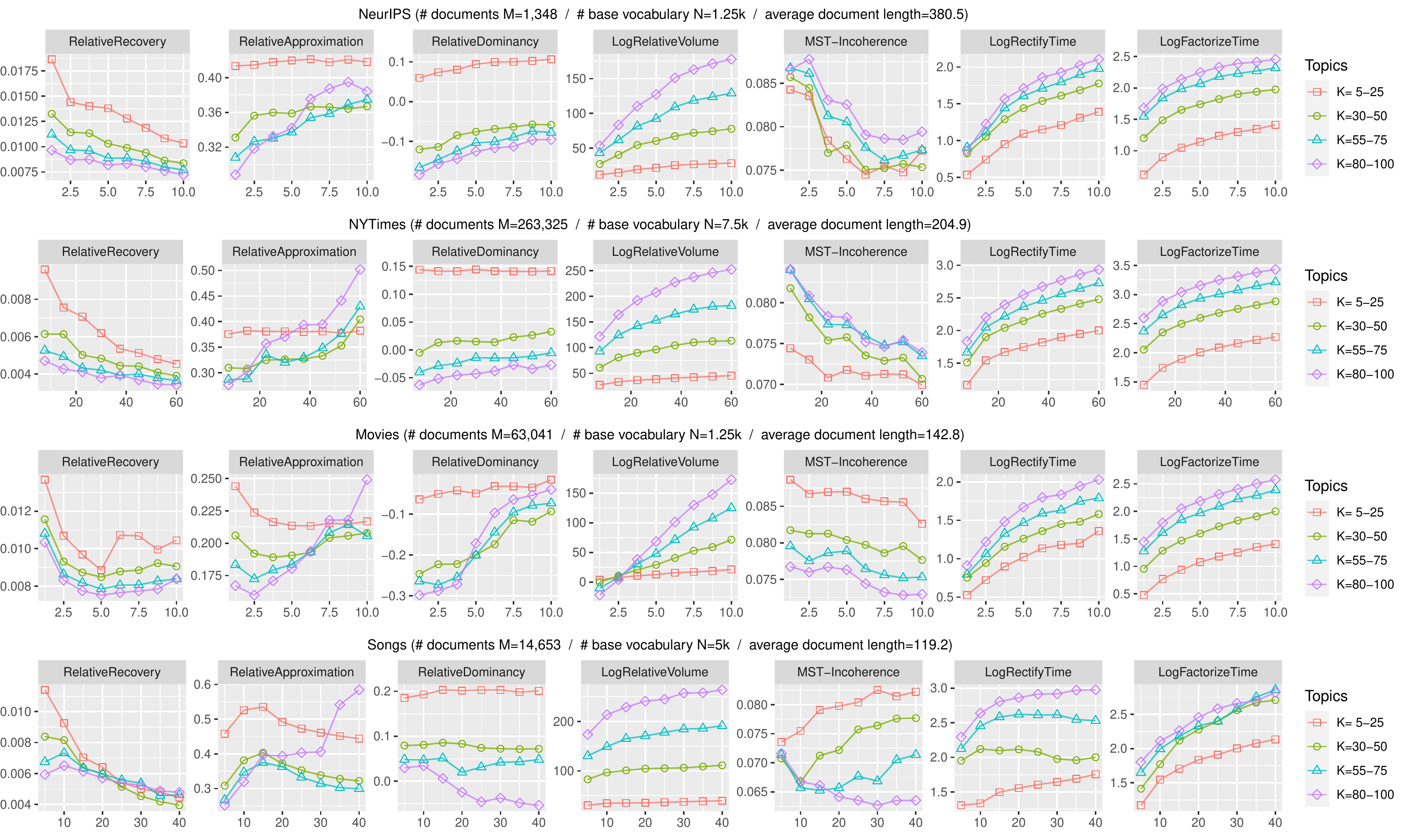}
	\vspace{-17px}
	\caption{As we increase the vocabulary size $N$, the relative volume of anchor convex hulls grows. MST-Incoherence decreases in general, but runtimes are stable. The $x$-axis shows $N$ in thousands. Vocabularies above 15k do not fit in memory on standard hardware with previous algorithms.}
	\label{fig:results-vocabs}
\end{figure*}

Figure \ref{fig:results-topics} shows the overall performance of the learned topics from the four datasets with increasing number of topics $K$. For fair comparisons across increasing vocabulary sizes later, we extend the intrinsic quality metrics used in \citep{moontae2015nips} to be relative to vocabulary scales. Low \textbf{RelativeRecovery} error \smash{ $(\frac{1}{N}\sum_i\|\BCbar_{i\ast} \Minus \BBreve_{i\ast}\BCbar_{\BS\ast}\|_2/\|\BCbar\|_F$)} implies that the learned anchor words successfully reconstruct the co\hyp{}occurrence space (as in Figure \ref{fig:cspace}) of the entire words. Low \textbf{RelativeApproximation} error $(\|\BC \Minus \BB\BA\BB^\transpose\|_F/\|\BC\|_F)$ means that our factorization captures most of information given in the unbiased co\hyp{}occurrence statistics. Topics in real data often exhibit correlations, and low \textbf{RelativeDominancy} $(\frac{1}{K}\sum_k \BA_{kk}/\|\BA\|_F)$ implies that our models learn more correlations between different topics. High \textbf{Specificity} \smash{$(\frac{1}{K}\sum_k\text{KL}(\BB_{\ast k}||\sum_i\BC_{\ast i}))$} indicates that the learned topics are distinct from the corpus unigram distribution, and high \textbf{Dissimilarity} tells that most top words in each topic do not occur within the top $20$ words of the other topics, showing interpretable difference across the learned topics. We do not report a traditional topic Coherence \citep{chang2009reading,mimno2011optimizing} as it often measures deceptively if a model learns many duplicated topics whose top words are mostly high-frequency unigrams \citep{huang2016}.

The first five columns show that using ENN+LAW or LR\hyp{}JSMF learn approximately same topics as using AP+AW without any visible loss in accuracy across all settings. More importantly, the randomness in LR\hyp{}JSMF produces very low variance over a number of runs. This is important as the stability of spectral inference is a major advantage over MCMC or Variational Inference. Although using PALM+LAW deviates slightly from the other three methods, it mostly achieves the same level of accuracy and follows the overall trend closely. Note that all of our methods have clear advantage over AP+AW, gaining $1\sim2$ orders of magnitude speedup in most situations. Even with the relatively small vocabularies, our algorithms show notable improvements in efficiency.

Despite the success of anchor-based topic modeling, the anchor words themselves are too rare terms to help interpretation of topics. In addition to the traditional \textit{Prominent Words (PWs)} selected from $\BB_{*k}$ for each topic $k$, we define \textit{Characteristic Words (CWs)} as the most co\hyp{}occurring terms with the anchor word $s_k$ with respect to \smash{$\BCbar_{s_k*}$}. As shown in Table \ref{tab:NIPStopics}, reading \textit{biological} (never chosen as a CW in the smaller vocabularies) clarifies that the first topic is more about neuroscience than computer science. Similarly reading \{\textit{character, kanji, radical}\} together with the PWs hints that the second topic describes written/spoken language recognition. By reading PWs and CWs altogether, users can better understand both general and specific details of the topic, also inspiring a new coherence metric.

\begin{table*}
\caption{5 topics from NeurIPS with vocabulary of size 10k. Left column: each line shows the top 6 words that contribute the most to the topic $k$ in $\BB_{*k}$. Right column: each line shows the top 6 words that co-occur most frequently with the anchor word $s_k$ in $\BCbar_{s_k*}$. Using characteristic words in bold in addition to prominent words enables more specific and definitive interpretation of topics.}
    \begin{center}
    \begin{tabular}{c c}
    \toprule
        {\bf \small{Top Prominent Words from $\BB$ by LR-JSMF}} & {\bf \small{Top Characteristic Words from $\BCbar$}}\\
        \midrule
        \small neuron cell circuit synaptic layer signal & \small neuron synaptic \textbf{potential} cell \textbf{biological} circuit\\
        \small layer recognition hidden word speech net & \small \textbf{character} \textbf{samples} \textbf{kanji} recognition \textbf{radical} layer\\
        \small control action dynamic optimal policy reinforcement & \small \textbf{tpdp} control \textbf{states} optimal action dynamic\\
        \small cell field visual image motion direction & \small motion cell direction \textbf{contrast} \textbf{signal} \textbf{region}\\
        \small gaussian noise approximation bound hidden matrix & \small \textbf{conditional} bound \textbf{likelihood} \textbf{cem} \textbf{log} gaussian\\
    \bottomrule
    \end{tabular}
    \end{center}
    \label{tab:NIPStopics}
    \vspace{-9px}
\end{table*}

For the second series of experiments, we create corpora $\{\BH_{kN}\}_{k=1}^8$ by
choosing vocabularies of $kN$ words with greatest tf-idf scores.
Here we do not compare LR-JSMF to ENN/PALM/AP as we cannot store full co\hyp{}occurrence matrices. 
Grouping results from $K \Eq 5$ to $100$ into four categories\footnote{For example, K=30-50 in Figure \ref{fig:results-vocabs} means that individual metrics are averaged over multiple runs using K=30, 35, 40, 45, and 50. It is to evaluate mean performance over varying ranges of $K$.}, Figure \ref{fig:results-vocabs} shows the 
performance of the 
topics with increasing vocabulary size.
High \textbf{LogRelativeVolume} means that the volume of an anchor convex hull (as in Figure \ref{fig:cspace}) is large relative to the average volume of random convex hulls in the same $N$-dimensional space. We measure incoherence of each topic as the minimum spanning tree cost of the associated graph where nodes are the union of 7 PWs and 7 CWs. Every node pair $(i, j)$ on the graph is linked with an undirected edge with weight $\frac{1}{2}(1-\text{NPMI}(i, j))$ where $\text{NPMI}(i, j) \Eq -\text{PMI}(i, j)/\log \BC_{ij}$ as in \citep{roder2015exploring}. Low \textbf{MST-Incoherence} implies that every topic has at least one path of top words that allows coherent understanding. When a topic consists of two relevant subtopics bridged by a few top words, our MST-Incoherence is less sensitive to possibly large pairwise distances between the top words of the two subtopics.\footnote{The graphs in MST-Incoherence are complete graphs of limited size (max 14 nodes), regardless of the vocab size. Adding more words does not allow more paths. Thus the metric relies solely on the associations amongst the prominent and characteristic words.}

As $N$ increases, the relative volumes of anchor convex hulls grow, while relative
recovery errors decrease.
This means inference quality
improves because LR-JSMF chooses better anchor words from larger vocabularies, thereby better representing non-anchor words inside the convex hulls. As each anchor word corresponds to a vertex of the growing anchor convex hulls, users are provided with more informative characteristic words over increasing $N$. The decreasing MST-Incoherence supports our intuition, again emphasizing the power of using large vocabulary. Most excitingly, the running times of ENN and LAW show the scalability of our new rectification and decomposition algorithms, thereby demonstrating the efficient and robust pipeline of LR-JSMF. 
Figure \ref{fig:results-vocabs} does not report Specificity and Dissimilarity because we cannot directly compare distributional distances measured on the different supports. 
The supplementary material includes all the missing panels in Figures \ref{fig:results-topics} and \ref{fig:results-vocabs}.

\section{Conclusion}
Spectral algorithms provide appealing alternatives for identifying interpretable low-rank subspaces by simple factorizations of higher-order co\hyp{}occurrence data. But this simplicity is also a weakness: 
the size of the co\hyp{}occurrence limit us to small vocabularies, and these methods perform poorly without rectifications that previously suffered quadratic scaling. Anchor words are guaranteed to be exclusive to the corresponding topics, but they are rarely used for topic interpretations because they are often chosen as too rare terms.


We develop a robust and scalable pipeline: Low-Rank Joint Stochastic Matrix Factorization based on our two complementary on-the-fly rectification methods (ENN/PALM) and a sufficiently general low-rank inference algorithm (LAW). These methods simultaneously compress and rectify the co\hyp{}occurrence from raw data; learn high-quality topics from the compressed matrix factorization; and achieve low-rank non-negative approximations without quadratic blowup. They also provide orders of magnitude speedups for rectification even on small vocabularies. In addition, we verify that using large vocabularies benefits inference quality by better satisfying the separability assumption. It also improves model interpretability by jointly understanding the prominent words with the characteristic words, and by measuring our MST-Incoherence metric for individual topics. Given all these new development, we can now learn and evaluate useful low-dimensional structures in high-dimensional datasets on laptop-grade hardware, massively increasing the applicability and potential use of the spectral algorithms.\footnote{The \textbf{supplementary material} of this paper consists of various friendly answers to potential questions from readers. Reading it together with the main paper will boost your understanding.}

\section*{Acknowledgements}
This work was designed when the first author was visiting Microsoft Research at Redmond. We thank both the department of Information and Decision Sciences in UIC Business School and the Deep Learning group in Microsoft Research. They provided useful resources and thoughtful ideas.

\bibliography{references}
\bibliographystyle{icml2021}

\end{document}